\newcommand{\CLASSINPUTtoptextmargin}{54pt}
\newcommand{\CLASSINPUTbottomtextmargin}{72pt}
\newcommand{\CLASSINPUTinnersidemargin}{45pt}
\newcommand{\CLASSINPUToutersidemargin}{45pt}

\documentclass[a4paper,conference]{IEEEtran}

\usepackage{amsmath,amssymb,amsfonts}
\usepackage{algorithmic}
\usepackage{graphicx}
\usepackage{textcomp}
\usepackage{hyperref}
\usepackage{xcolor}
\usepackage{soul}
\usepackage{tikz}

\sethlcolor{red!30}

\usepackage[acronym]{glossaries}
\makeglossaries

\newacronym{llm}{LLM}{Large Language Model}
\newacronym{osm}{OSM}{OpenStreetMap}
\newacronym{gui}{GUI}{Graphical User Interface}
\newacronym{vlm}{VLM}{Vision Language Model}
\newacronym{sota}{SotA}{State-of-the-Art}
\newacronym{vqa}{VQA}{Visual Question Answering}
\newacronym{mmr}{MMR}{Maximum Marginal Relevance}

\def\BibTeX{{\rm B\kern-.05em{\sc i\kern-.025em b}\kern-.08em
T\kern-.1667em\lower.7ex\hbox{E}\kern-.125emX}}

\newcommand\submittedtext{%
	\footnotesize This work has been submitted to the IEEE for possible publication. Copyright may be transferred without notice, after which this version on arXiv will be replaced with the Version of Record.}

\newcommand\submittednotice{%
	\begin{tikzpicture}[remember picture,overlay]
		\node[anchor=south,yshift=10pt] at (current page.south) {\fbox{\parbox{\dimexpr0.65\textwidth-\fboxsep-\fboxrule\relax}{\submittedtext}}};
	\end{tikzpicture}%
}

\begin{document}
    \bstctlcite{IEEEexample:BSTcontrol}
    \title{Multimodal Spatio-Temporal Summarization of Localized Image Sequences\\
    }

    \author{
        \IEEEauthorblockN{\begin{tabular}{cc}
                              1\textsuperscript{st} Nicolas Schuler\IEEEauthorrefmark{1}\IEEEauthorrefmark{2} &
                              2\textsuperscript{nd} Janik Kurtz\IEEEauthorrefmark{1} \\
                              \href{https://orcid.org/0009-0007-4098-1244}{{\includegraphics[keepaspectratio,width=0.7em]{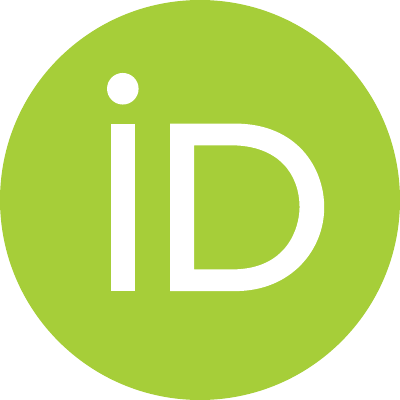}}} 0009-0007-4098-1244 &
                              \href{https://orcid.org/0009-0003-6792-2245}{{\includegraphics[keepaspectratio,width=0.7em]{images/ORCID_iD}}} 0009-0003-6792-2245
        \end{tabular}}
        \\
        \IEEEauthorblockN{\begin{tabular}{cc}
                              3\textsuperscript{rd} Félicia Teferle\IEEEauthorrefmark{2} &
                              4\textsuperscript{th} Jürgen Graf\IEEEauthorrefmark{1} \\
                              \href{https://orcid.org/0000-0002-9132-3571}{{\includegraphics[keepaspectratio,width=0.7em]{images/ORCID_iD}}} 0000-0002-9132-3571 &
                              \href{https://orcid.org/0000-0002-1354-0888}{{\includegraphics[keepaspectratio,width=0.7em]{images/ORCID_iD}}} 0000-0002-1354-0888
        \end{tabular}}
        \\
        \IEEEauthorblockA{\begin{tabular}{cc}
                              \IEEEauthorrefmark{1} \textit{Department of Computer Science} &
                              \IEEEauthorrefmark{2} \textit{Department of Engineering} \\
                              \textit{Trier University of Applied Sciences} &
                              \textit{University of Luxembourg} \\
                              Trier, Germany &
                              Kirchberg, Luxembourg
        \end{tabular}
        }
    }

    \maketitle

    \begin{abstract}
        Summarization and highlight detection are essential for processing the
        expanding volume of multimodal data in robotics and autonomous
        navigation.
        While recent methods rely on pre-trained embeddings and transformer
        architectures via end-to-end learning or fixed heuristics, these can be
        difficult to adapt to domain-specific requirements or user-specific
        selection preferences.
        This paper introduces a modular framework that integrates visual and
        textual embeddings with temporal and geospatial metadata through
        individual scoring functions, enabling customizable keyframe extraction
        and visual summarization of localized recordings.
        We evaluate the proposed approach on a custom dataset comprising a
        single trajectory and show that different weight configurations
        yield different frame selections, illustrating how the proposed
        subscores influence characteristics of the generated visual summary
        within this experimental setting.
        A perturbation-based sensitivity analysis further characterizes the
        response of the frame selection process to changes in individual
        weights and reveals broadly consistent sensitivity patterns across the
        examined perturbation magnitudes.
        Code, example videos and images, anonymized example data, evaluation metrics,
        and experimental results are available at
        \href{https://datahub.rz.rptu.de/hstr-csrl-public/publications/multimodal-spatio-temporal-summarization}
        {https://datahub.rz.rptu.de/hstr-csrl-public/publications/multimodal-spatio-temporal-summarization}.

    \end{abstract}


	\submittednotice

    \section{Introduction}\label{sec:introduction}
    The rapid growth of visual and multimodal sensor data in scene
    interpretation, robotics, and navigation necessitates computing concise
    representations and selecting representative data subsets for downstream
    tasks~\cite{Kawaharazuka2025,NVIDIA2026,Lee2025}.

    While modern approaches for summarization and keyframe extraction
    using \glspl*{llm}~\cite{Lee2025} and \glspl*{vlm}~\cite{NVIDIA2026} succeed
    in embedding multimodal data into pre-trained representation
    spaces~\cite{Simeoni2025}, their mostly end-to-end architectures limit
    explainability and user- or domain-specific control~\cite{Mumuni2025}.
    In contrast, two-stage approaches can provide greater interpretability by
    pairing pre-trained embedding models with explicit scoring
    functions~\cite{Chen2026a}.

    The present work proposes a modular framework that combines pre-trained
    multimodal embeddings with custom scoring metrics to extract representative
    keyframes of localized image sequences.
    By utilizing visual and textual embeddings with temporal and geospatial
    metadata, our approach allows users to prioritize features tailored to
    their specific applications.

    \section{Foundations and Related Work}\label{sec:foundations-and-related-work}
    This section reviews the technical foundations and related work relevant to
    the proposed approach, including embedding-based similarity, video
    summarization, and report generation in mobile robotics.

    \subsection{Embedding, Similarity and Summarization}\label{subsec:embedding-similarity-and-summarization}
    Embedding various data modalities, including visual, audio and textual
    data, into high-fidelity representation spaces has become a key challenge
    to successfully employ these representations in various downstream
    tasks~\cite{Radford2021, Girdhar2023}.
    Among these tasks are zero-shot object detection and
    segmentation~\cite{Simeoni2025}, similarity scoring~\cite{Gupta2025},
    text generation~\cite{Cao2024} and \gls*{vqa}~\cite{NVIDIA2026}.

    The embedding process maps the input into a vector or matrix that captures
    its underlying semantic relationships~\cite{Zhang2025a}.
    Popular embedding models include DINOv3~\cite{Simeoni2025} for image
    embedding, CLIP~\cite{Radford2021} for shared text and image embedding and
    SBERT~\cite{Reimers2019} for text embedding, with
    \glspl*{llm} and \glspl*{vlm} typically integrating the embedding process
    directly into their end-to-end
    architecture~\cite{OpenAI2025, Adcock2026TheL4}.
    These representations can then be used to compute data similarity, either
    using an explicit metric such as cosine similarity or through an
    end-to-end model, e.g.\ within a transformer architecture~\cite{Gupta2025}.
    These similarity scores serve to summarize data, enabling tasks such as
    keyframe extraction from video sequences~\cite{Pang2025} and multimodal
    summarization~\cite{He2024}.

    Building on this, in computer vision and robotics, automated report
    generation can be defined as the automated extraction of visual, action,
    trajectory, and multimodal sensor data into structured natural language
    summaries~\cite{DeChant2023}.
    This task is of particular interest in industrial~\cite{Gao2025} and medical
    inspection~\cite{Yuan2025} as well as autonomous robotics~\cite{DeChant2023}.
    All share the common challenge of processing large, heterogeneous data
    streams which are often obscure and inaccessible to laypersons, hindering
    efficient supervision and evaluation.

    \subsection{Related Work}\label{subsec:related-work}
    For summarization and report generation, we differentiate between two
    paradigms: end-to-end and two-stage~\cite{Chen2026a, Lee2025}.
    Unlike end-to-end video summarization models, two-stage approaches first
    extract dense representations using previously introduced embedding models.
    Subsequently, they apply explicit, manually defined scoring functions to
    score and select keyframes.
    These include graph centrality~\cite{Sahu2023}, \gls*{mmr}~\cite{Chen2026a},
    or pairwise embedding similarities~\cite{Tan2024}.

    Our approach falls into the latter category, applying distinct scoring
    functions to enhance both customizability and explainability.
    Similar to our approach, AKS~\cite{Tang2025} combines relevance and temporal
    coverage in its keyframe selection process, while~\cite{Lee2025} uses a
    similar approach to filter out spatio-temporal redundancies.
    ~\cite{Chen2026a} utilizes DINOv2 embeddings with relevance scoring and
    redundancy penalties to select keyframes.
    Our method is distinct in its integration of diverse scoring elements into
    a unified, user-accessible framework, which is designed to support system
    explainability and data traceability.

    \section{Multimodal Spatio-Temporal Summarization}\label{sec:multimodal-spatio-temporal-summarization}
    This section introduces the proposed multimodal spatio-temporal
    summarization approach, including the underlying agents and sensory systems
    as well as the proposed pipeline for generating user-specific visual
    summaries from image sequences.

    \subsection{Platforms and Sensory Systems}\label{subsec:platforms-and-sensory-systems}
\begin{figure}[!htbp]
        \vspace{5pt}
        \centering
        \includegraphics[width=0.95\columnwidth]{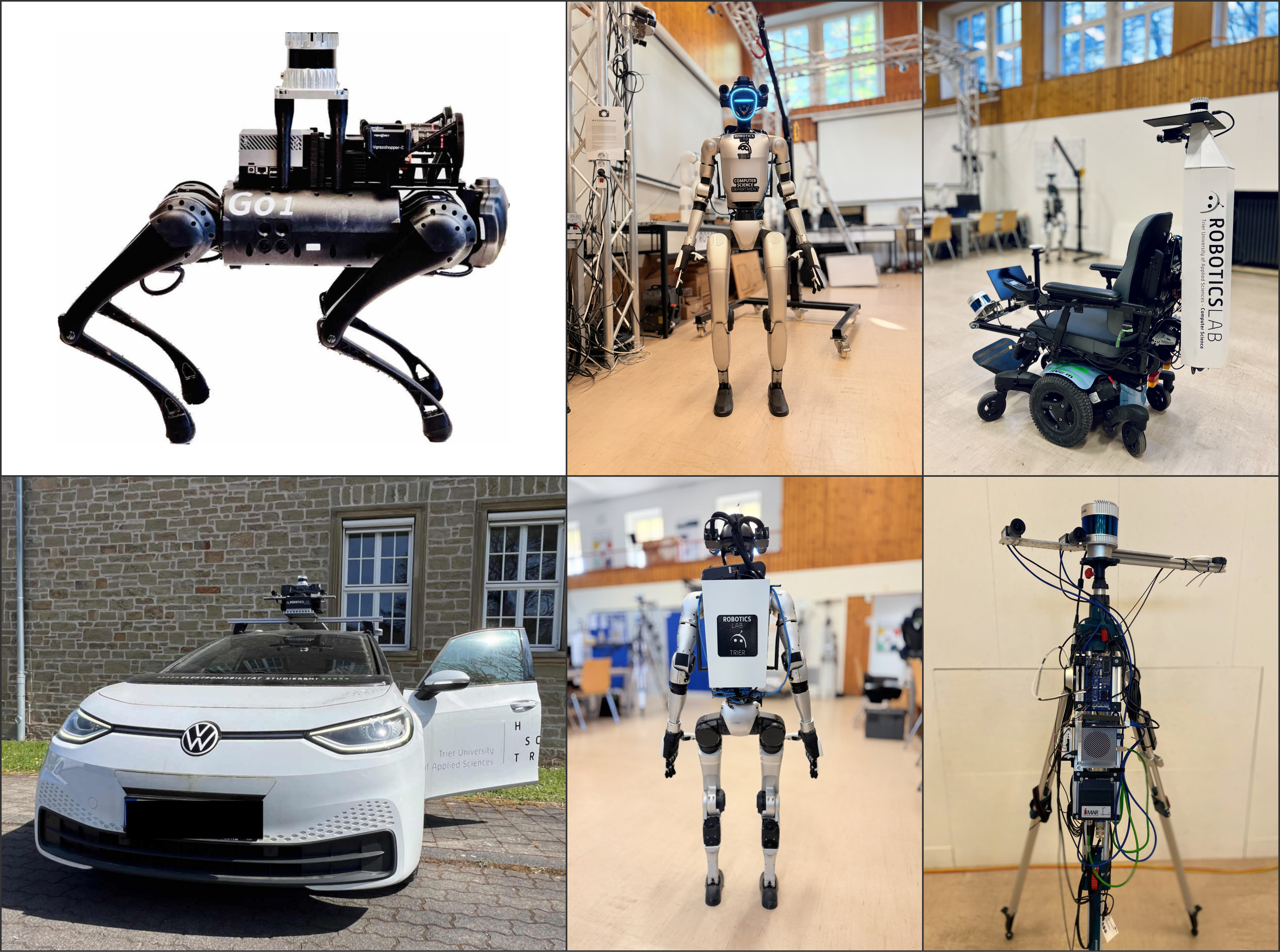}
        \caption{
            \textbf{Different cognitive systems used within our laboratory}.
            Unitree Go1 (top left), vehicle (bottom left), Unitree G1
            (top and bottom mid), wheelchair (top right) and tripod (bottom
            right) with custom modular sensory system setups.
            Two images were previously published in~\cite{Schuler2025}, while
            the remaining images and updated descriptions were introduced
            in~\cite{Schuler2026}.
        }
        \label{fig:fig1}
    \end{figure}

    We detail the different mobile and static platforms used within our laboratory.
    Our agents include humanoid and quadruped robots (Unitree G1 and Go1), various
    types of electric wheelchairs, a research vehicle (VW ID.3) and multiple
    tripods (Bosch BT 300 HD).
    Fig.~\ref{fig:fig1} depicts some of these agents with custom sensory
    setups.
    These setups leverage the specific requirements for each agent.
    Shared configurations include stereo RGB video cameras (FLIR FG-P5G-50S4C-C),
    3D LiDAR (Ouster OS1-64 Rev.\ 7, HESAI Pandar 64, LIVOX MID-360), INS
    (iMAR iNAT) and GNSS (u-blox F9R).
    We deploy a custom software stack on an NVIDIA Jetson AGX Orin in combination
    with a custom circuit board to record time-synchronized data.
    We utilize this system stack to record a multimodal pilot dataset to
    evaluate the proposed approach.

    \subsection{Proposed Pipeline}\label{subsec:proposed-pipeline}
    This section introduces the proposed pipeline to generate customized
    summarizations of image sequences, resulting in a visual summary.
    Note that this serves as an overview of the approach.
    Details regarding the actual implementation, models used and algorithms are
    discussed in Section~\ref{subsec:experimental-setup}.

    The core idea of the approach is to select images from a recorded,
    localized image sequence which can serve to summarize this sequence.
    These selected image frames form a visual summary.
    Unlike automatic highlight detection, the objective is to generate a visual
    summary tailored to the interests of an individual user.
    To that end, the user configures a set of seven weights $w_1, \dots, w_7$
    to balance the corresponding subscores that contribute to the final score $F$
    to select the relevant images $\mathcal S$ from the sequence containing all
    images $\mathcal I$.
    We define these subscores  $G_{\text{vis}}$, $G_{\text{txt}}$, and
    $G_{\text{enr}}$ as Marginal Log-Determinant Diversity Gains for visual,
    textual, and enriched textual semantics, respectively, alongside
    Visual Semantic Novelty $N_{\text{vis}}$, Geometric Relevance
    $R_{\text{geo}}$, Temporal Distribution $D_{\text{tmp}}$
    and Spatial Distribution $D_{\text{spt}}$.
    Below, we outline the conceptual foundations of these subscores.
    The exact implementation and mathematical definitions are provided in
    Section~\ref{subsec:experimental-setup}.

    The subscores $G_{\text{vis}}$, $G_{\text{txt}}$, and $G_{\text{enr}}$ share a
    unified mathematical foundation, differing solely in their underlying embedding
    spaces.
    The objective of these metrics is to promote diversity of embeddings upon
    the insertion of a new candidate $c$ into $\mathcal{S}$.
    Specifically, $G_{\text{vis}}$ operates directly on the visual image data,
    whereas $G_{\text{txt}}$ and $G_{\text{enr}}$ leverage natural language
    descriptions generated from those frames.
    The distinction between $G_{\text{txt}}$ and $G_{\text{enr}}$ lies exclusively
    in the granularity of the textual descriptions provided.
    The image data are acquired by the sensory system, while the descriptions
    are generated using \glspl*{vlm}.
    To extract meaningful embeddings from the raw data, architectures like
    DINOv3~\cite{Simeoni2025} for visual feature extraction,
    CLIP~\cite{Radford2021} for joint vision-language representations and
    sentence transformers~\cite{Reimers2019} for textual embeddings can be
    employed.

    The subscores $D_{\text{tmp}}$ and $D_{\text{spt}}$ promote
    the temporal and spatial information coverage by scoring each candidate $c$
    based on its proximity to all elements of $\mathcal S$.
    To achieve this, each image frame is mapped to its absolute geospatial
    coordinates and corresponding synchronized timestamp.

    While these five subscores are calculated iteratively for each new candidate
    $c$ being added to $\mathcal S$, the remaining two subscores,
    $N_{\text{vis}}$ and $R_{\text{geo}}$, are calculated once for each
    candidate, independent of the composition of $\mathcal S$.
    For each image candidate $c$, $N_{\text{vis}}$ is computed relative to its
    surrounding frames within the temporal radius $\tau$, where the subscore is
    defined as the mean dissimilarity between $c$ and its neighboring frames.
    The subscore leverages the identical embedding space utilized by
    $G_{\text{vis}}$.
    Given a set of predefined geospatial entities $\mathcal O$, subscore
    $R_{\text{geo}}$ evaluates each image frame based on its proximity to these
    objects of interest.
    The subscore is based on the distance to the nearest target geospatial
    feature, assigning higher scores to candidates closer to relevant objects.

    Let all subscores be implicitly evaluated for the candidate $c$
    and, where applicable, conditioned on the current set of selected images
    $\mathcal S$ and the set of objects $\mathcal O$.
    The total candidate-specific score $F$ for a candidate image $c$ is defined
    as the linear combination of its individual subscores:

    \begin{equation}
        \begin{split}
            F(c\mid\mathcal S,\mathcal O) = {} & w_1 G_{\text{vis}} + w_2 G_{\text{txt}} + w_3 G_{\text{enr}} + w_4 N_{\text{vis}} \\
            & + w_5 R_{\text{geo}} + w_6 D_{\text{tmp}} + w_7 D_{\text{spt}}.
        \end{split}
    \end{equation}

    Thus, the image candidate $c \in \mathcal I \setminus \mathcal S$ with the
    highest score $F$ is added to the set of representative images
    $\mathcal S$.
    Before calculating the final score $F$, the individual subscores for every
    candidate $c$ are normalized across all candidates once per iteration to
    prevent one subscore dominating the selection process under identical
    weighting parameters.
    We iteratively add candidates to $\mathcal S$ until a pre-defined
    cardinality limit $K$ is reached, i.e. $|\mathcal S| = K$.

    \section{Evaluation}\label{sec:evaluation}
    This section presents the experimental evaluation of the proposed approach,
    including the experimental setup, results, and a discussion of its
    strengths and limitations.

    \subsection{Experimental Setup}\label{subsec:experimental-setup}
    To evaluate the proposed approach, we utilize data recorded on a wheelchair
    platform at Trier University of Applied Sciences.
    For an illustration of the setup used, refer to Fig.~\ref{fig:fig1}.
    The data include images from RGB stereo cameras (FLIR FG-P5G-50S4C-C),
    recorded at 20 Hz, 3D point clouds from a LiDAR (Ouster OS1-64 Rev.\ 7),
    recorded at 10 Hz, and pose information from an INS (iMAR iNAT), recorded
    at 100 Hz.
    All data are temporally synchronized and geo-referenced.
    We do not evaluate our approach on commonly used video summarization
    datasets such as SumMe~\cite{Gygli2014} and TVSum~\cite{Song2015}, as these
    datasets do not capture the characteristics required by our proposed
    framework.
    Our approach is designed to leverage temporally synchronized and
    geospatially localized multimodal recordings not available in standard
    video summarization benchmarks, making them unsuitable to evaluate our
    approach.

    In addition, we calculate the pose of the recording agent using scan-to-map
    matching~\cite{Olson2009} by matching the local point clouds obtained from
    the LiDAR to a pre-scanned map of the campus at Trier University of Applied
    Sciences.
    The final recording used consists of 8,464 frames.
    For the purpose of this paper, we only utilize the scan-to-map matching poses
    and the images of the left camera.

    The pipeline is implemented using Python 3.13, PyTorch 2.11 and CUDA 12.6.
    Exemplary inference runs were performed on a workstation equipped with an
    AMD Ryzen 5 5600X CPU and an NVIDIA GeForce RTX 4070 Ti SUPER GPU\@.
    The sensitivity analysis was performed on an HPC cluster at RPTU
    Kaiserslautern-Landau using multiple CPU configurations and the Slurm
    workload manager.

    To score all candidate images, we apply the seven subscores previously
    introduced: $G_{\text{vis}}$, $G_{\text{txt}}$, $G_{\text{enr}}$,
    $N_{\text{vis}}$, $R_{\text{geo}}$, $D_{\text{tmp}}$ and $D_{\text{spt}}$.
    If multiple candidates tie for the combined score $F$, the candidate
    with the earliest timestamp is added to $\mathcal S$.

    First, we describe the scorings $G_{\text{vis}}$, $G_{\text{txt}}$ and
    $G_{\text{enr}}$.
    We generate action descriptions of the recording using
    Cosmos-Reason2-8B~\cite{NVIDIA2025, NVIDIA2025a} at three-second intervals.
    The 8,464 frames are partitioned into 141 complete batches of 60 frames and
    one final batch containing four frames, resulting in 142 descriptions.
    We provide the prompt used to generate the descriptions in the
    \href{https://datahub.rz.rptu.de/hstr-csrl-public/publications/multimodal-spatio-temporal-summarization}
    {external repository}.
    We augment these descriptions with additional information including
    person reidentification and geospatial data like building names and surface
    composition taken from \gls*{osm}~\cite{OSMF}.
    This yields two textual representations for each batch: the original action
    description used for $G_{\mathrm{txt}}$, and an enriched description
    incorporating person re-identification and geospatial information, used
    for $G_{\mathrm{enr}}$.
    We detail this approach in another publication~\cite{Schuler2026}.
    For the purpose of this paper, we utilize these augmented descriptions to
    offer an alternative, more diverse description to embed.
    The images are embedded via DINOv3~\cite{Simeoni2025} ViT-B16 using the
    Transformers~\cite{Wolf2019} library, the descriptions are embedded via
    MPNet~\cite{Song2020} Base v2 using SentenceTransformers~\cite{Reimers2019}.
    The embeddings are then normalized.
    To calculate the individual subscores for all image candidates given the
    embeddings described above, we compute a marginal log-determinant
    gain~\cite{Guestrin2005}, $g$, for each candidate $c$ to be added to the
    set of representative images $\mathcal S$, defined as

    \begin{equation}
        g(x_c\mid X_{\mathcal S})=
        \begin{cases}
            0, & |\mathcal S|=0,\\[2mm]
            \log\!\left(1-\left\|L^{-1}X_{\mathcal S}x_c\right\|_2^2\right),
            & |\mathcal S|>0,
        \end{cases}
    \end{equation}

    where $x_c\in\mathbb{R}^{d}$ is the normalized feature vector of
    candidate $c$ and $X_{\mathcal S}\in\mathbb{R}^{|\mathcal S|\times d}$
    contains the normalized feature vectors of the selected representative
    images with an embedding size of $d$.
    Further, $L\in\mathbb{R}^{|\mathcal S|\times|\mathcal S|}$ denotes the
    Cholesky factor of the Gram matrix $K$ associated with the selected set
    $\mathcal S$:

    \[
        K_{\mathcal S}=X_{\mathcal S}X_{\mathcal S}^\top=LL^\top.
    \]

    Because all feature vectors are $\ell_2$-normalized, the entries of
    $X_{\mathcal{S}}x_c^\top$ correspond to cosine similarities between the
    embeddings of the candidate and the selected images.
    The term $1-\left\|L^{-1}X_{\mathcal S}x_c^{\top}\right\|_2^2$ represents
    the residual term corresponding to the additional diversity provided by
    candidate $c$.
    This incremental Cholesky scheme is described in~\cite{Chen2017}.
    For numerical stability, this quantity is clipped to a minimum value
    $\delta>0$ before applying the logarithm, i.e.,

    \[
        \max\!\left(
                  1-\left\|L^{-1}X_{\mathcal S}x_c^{\top}\right\|_2^2,
                  \delta
        \right),
        \qquad
        \delta = 10^{-12}.
    \]

    The individual log-determinant gain subscores are then expressed by mapping
    their respective embedding spaces to this formulation:
    $G_{\text{vis}}(c\mid\mathcal S) = g(x_c^{\text{vis}}\mid X_{\mathcal{S}}^{\text{vis}})$,
    $G_{\text{txt}}(c\mid\mathcal S) = g(x_c^{\text{txt}}\mid X_{\mathcal{S}}^{\text{txt}})$,
    and $G_{\text{enr}}(c\mid\mathcal S) = g(x_c^{\text{enr}}\mid X_{\mathcal{S}}^{\text{enr}})$.
    Since the used recording contains 8,464 frames but only 142 descriptions, each
    batch of 60 consecutive frames is assigned the same description when
    calculating the subscores for individual frames.

    Unlike the diversity gain metrics above, the Visual Semantic Novelty
    $N_{\text{vis}}$ is independent of the already selected set of representative
    images $\mathcal{S}$, allowing it to be pre-computed once for the entire
    sequence.
    Let $c_t$ denote the image candidate at frame index $t$ and let $x_t$ be
    its corresponding normalized DINOv3 image embedding.
    We define $N_{\text{vis}}(c_t)$ as the mean cosine distance between the
    candidate and its surrounding frames within a symmetric temporal window

    \begin{equation}
        N_{\text{vis}}(c_t) = \frac{1}{|\mathcal{W}_t|} \sum_{i \in \mathcal{W}_t} \left( 1 - x_t^{\top}x_i \right),
    \end{equation}

    where $\mathcal{W}_t = \{ i \in \mathbb{N} \mid t - \tau \le i \le t + \tau, \, i \ne t \}$
    represents the set of neighboring frame indices.
    The window radius is set to $\tau = 30$ frames to symmetrically mirror the
    60-frame sequence length utilized during the natural language description
    generation process at the recorded 20 Hz frame rate.
    At sequence boundaries, $W_t$ is truncated to the available neighboring
    indices.

    Similarly, the subscore $R_{\text{geo}}$ is calculated once, independent of
    the selected candidate set $\mathcal S$.
    For each \gls*{osm} object of relevance, we calculate the distance in
    meters of the agent to that object for each image frame.
    For our dataset, we define objects of relevance as all buildings, paths,
    parking spaces, sports facilities and garbage bins within the area of the
    recording.
    We obtain the object data via the \gls*{osm} Overpass API and use the pose of
    the agent.
    Let $\mathcal O$ denote the set of user-selected relevant \gls*{osm}
    objects and let $d_{\mathrm{geo}}(c,o)$ denote the metric distance in
    meters between candidate $c$ and object $o \in \mathcal O$.
    The subscore is defined as the negative distance to the nearest relevant
    object

    \begin{equation}
        R_{\mathrm{geo}}(c\mid \mathcal O) =
        -\min_{o\in\mathcal O} d_{\mathrm{geo}}(c,o).
    \end{equation}

    Thus, frames closer to a relevant object receive higher scores.
    The user selects the relevant \gls*{osm} objects from the available objects
    through the provided interface.

    Both subscores $D_{\text{tmp}}$ and $D_{\text{spt}}$ are calculated
    given the already selected set of representative images $\mathcal S$.
    For each candidate, we calculate its distance to all previously selected
    images and use the minimum distance as its final value.
    This distance is the temporal and spatial distance in seconds and meters
    respectively and serves as the subscore value.
    Let $t_c$ denote the timestamp of candidate $c$ and $\mathbf{p}_c \in \mathbb{R}^3$
    its spatial position in ECEF coordinates.
    Given $\mathcal S$, the temporal and spatial distribution subscores are
    defined as

    \begin{equation}
        D_{\mathrm{tmp}}(c\mid\mathcal S)=
        \begin{cases}
            0, & |\mathcal S|=0,\\[2mm]
            \displaystyle\min_{s\in\mathcal S}|t_c-t_s|,
            & |\mathcal S|>0,
        \end{cases}
    \end{equation}

    and

    \begin{equation}
        D_{\mathrm{spt}}(c\mid\mathcal S)=
        \begin{cases}
            0, & |\mathcal S|=0,\\[2mm]
            \displaystyle\min_{s\in\mathcal S}
            \|\mathbf{p}_c-\mathbf{p}_s\|_2,
            & |\mathcal S|>0.
        \end{cases}
    \end{equation}

    respectively.
    Higher values therefore favor candidates that are farther from the already
    selected images in temporal or spatial space.
    Parallel to the embeddings above promoting the diversity gain via the
    embedding distances, these subscores encourage coverage of temporally and
    spatially separated portions of the sequence.

    By definition of $\mathcal S$, \ $G_{\text{vis}}$, $G_{\text{txt}}$,
    $G_{\text{enr}}$, $D_{\text{tmp}}$ and $D_{\text{spt}}$,
    these subscores return a constant value of $0$ should $\mathcal S$ be
    empty, that is on first iteration.
    Similarly, the subscore $R_{\text{geo}}$ may return identical top scores
    for multiple candidates, e.g.\ if multiple images are taken within an
    object of interest, reducing the distance and thus maximal score to $0$.
    If the weights of the remaining subscores are set to zero, a tie will
    occur.
    We resolve ties by adding the frame out of all top scoring candidates
    to the set that occurs first within the recording, i.e.\ that has the
    lowest index and thus timestamp.
    Before computing the final candidate score $F$, each subscore is
    standardized independently across the remaining candidate images.
    For a subscore $F_j \in \{G_{\text{vis}}, G_{\text{txt}}, G_{\text{enr}},
    N_{\text{vis}}, R_{\text{geo}}, D_{\text{tmp}}, D_{\text{spt}}\}$, its
    normalized value for candidate $c$ is given by

    \begin{equation}
        \widetilde{F}_j(c)
        =
        \frac{F_j(c) - \mu_j}
        {\sigma_j + \delta},
        \qquad
        \delta = 10^{-8},
    \end{equation}

    where \(\mu_j\) and \(\sigma_j\) denote the mean and standard deviation of
    subscore \(F_j\) across all remaining candidates.
    This standardization places the individual subscores on a comparable scale
    and prevents differences in their numerical ranges from causing one
    subscore to dominate the combined score.
    Thus, in the present implementation the weights operate on the normalized
    subscores defined above to calculate the final score $F$.

    To allow the user to adjust the weights associated with the subscores
    according to their needs, we implement a \gls*{gui} using Python.
    Besides the scoring weights, the user specifies a top-K integer value,
    which dictates the exact number of representative image frames to be
    selected, i.e.\ $|\mathcal S| = K$.
    When first using the program, all embeddings and subscores that do not have
    to be computed on the fly are calculated and saved locally.
    The program offers two additional functionalities via \gls*{llm}
    integration.
    First, the user can enter a free-text description of what they are
    interested in within the sequence.
    This description is then given as a prompt to an \gls*{llm}, which in turn
    suggests parameter values the user can utilize for their custom
    summarization.
    Secondly, the \gls*{gui} summarizes the descriptions associated with the
    selected representative images to a new description focusing on the
    selected set.
    We provide the prompts used in the
    \href{https://datahub.rz.rptu.de/hstr-csrl-public/publications/multimodal-spatio-temporal-summarization}
    {external repository}.
    We utilize gpt-oss-120b~\cite{OpenAI2025} within a local Open
    WebUI~\cite{Baek2025} instance as the \gls*{llm} agent running on a cluster
    with four NVIDIA H200 within the university network.

    We evaluate the implementation in two steps based on the test dataset
    described above.
    First, we perform a perturbation-based sensitivity analysis~\cite{Sobol2010}
    by varying the seven subscore weighting parameters previously introduced.
    Since the seven parameters are non-negative weights contributing to the final
    score $F$, they satisfy $w_i \geq 0$ and via normalization $\sum_{i=1}^{7}w_i=1$,
    restricting the parameter space to a six-dimensional simplex.
    By uniformly sampling weights from a Dirichlet distribution with
    concentration parameters \(\boldsymbol{\alpha}=\mathbf{1}\) and adding a
    perturbation \(\epsilon\) to one weight at a time, we evaluate the effect
    of the individual subscore weights on the selection of representative image
    frames.
    We renormalize the weights after adding $\epsilon$ to ensure that they remain
    on the simplex.

    Sensitivity is quantified using three complementary distances between the
    unperturbed and perturbed frame selections.
    When ordering two sets of equal cardinality containing the selected frame
    indices, their mean absolute difference is equivalent to the
    one-dimensional Wasserstein-1 distance $\Delta_{\mathrm{idx}}$ along the
    temporal (frame-index) axis.
    We additionally measure the temporal and spatial displacements,
    $\Delta_{\mathrm{tmp}}$ and $\Delta_{\mathrm{spt}}$, using the
    one-dimensional Wasserstein-1 distance between the corresponding timestamps
    and the three-dimensional Wasserstein-1 distance between the associated
    positions, respectively.

    Together, these metrics quantify how perturbations affect the temporal
    positions and physical locations of selected frames.
    We perform this analysis once for each
    $\epsilon \in \{0.001, 0.005, 0.01, 0.02, 0.05, 0.1, 0.25\}$ with a sample size
    $n=10{,}000$ over the simplex.
    Since each of the seven weights is perturbed once for every sampled
    weight configuration and for each of the seven perturbation magnitudes,
    the experiment comprises
    $10{,}000 \times 7 \times 7 = 490{,}000$
    perturbed image summarization selections, in addition to
    $10{,}000$ unperturbed selections.
    The sensitivity of the seven weighting parameters was evaluated across
    randomly sampled valid \gls*{osm} object configurations, with one to three
    available options activated per configuration.
    We set $|\mathcal S| = K = 5$.
    The goal of this analysis is twofold.
    First, to characterize the local sensitivity of the frame selection process
    to individual subscore weights.
    Second, to examine whether different subscores exhibit similar or
    limited sensitivity across the sampled weight configurations, potentially
    indicating overlapping or redundant contributions to frame selection.

    We empirically evaluate the implementation on the introduced
    dataset to assess its behavior and practical utility.
    We present representative outputs and investigate the effect of the
    different subscore weights $w_1, \dots, w_7$ on the resulting representative
    image frame selections.
    The evaluation aims to highlight both the strengths of the proposed
    approach and potential limitations.

    \subsection{Results}\label{subsec:results}
\begin{figure}[!htbp]
        \centering
        \includegraphics[width=0.95\columnwidth]{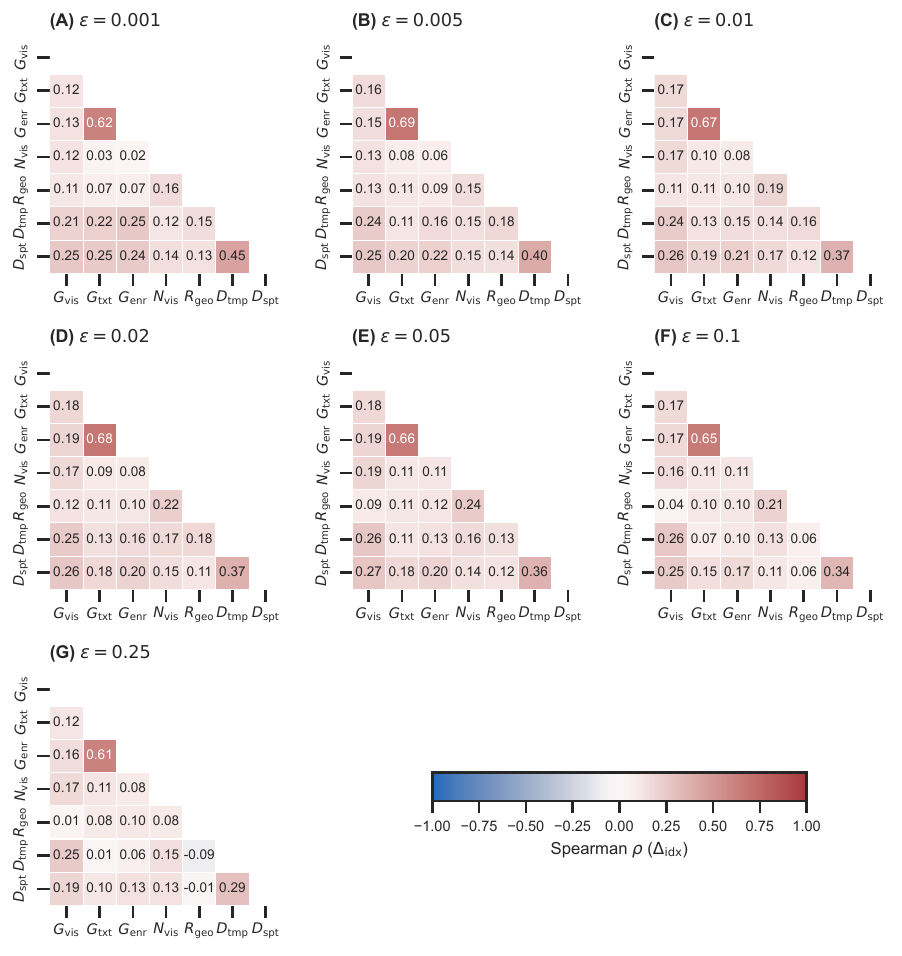}
        \caption{
            \textbf{Spearman correlations between weight-specific local
            sensitivities across sampled weight configurations}.
            Each heatmap shows the Spearman rank correlation between the
            normalized output sensitivities measured using $\Delta_{\text{idx}}$
            obtained by individually perturbing the proposed subscore weights.
            The correlations are computed across $n=10{,}000$ sampled base
            weight configurations.
        }
        \label{fig:fig2}
    \end{figure}

    Fig.~\ref{fig:fig2} shows the Spearman correlation between
    weight-specific local sensitivities measured using $\Delta_{\text{idx}}$.
    These correlations are between local sensitivities and do not imply a
    correlation between the corresponding subscore weights.
    The highest correlation is observed between the sensitivities of
    $G_{\text{txt}}$ and $G_{\text{enr}}$, ranging from $0.61$ to $0.69$ across
    the examined perturbation magnitudes $\epsilon$.
    In general, the observed correlation patterns remain stable over the
    examined perturbation magnitudes $\epsilon$, except for the largest
    perturbation, $\epsilon = 0.25$, shown in Subplot~(G).

\begin{figure}[!htbp]
        \centering
        \includegraphics[width=0.95\columnwidth]{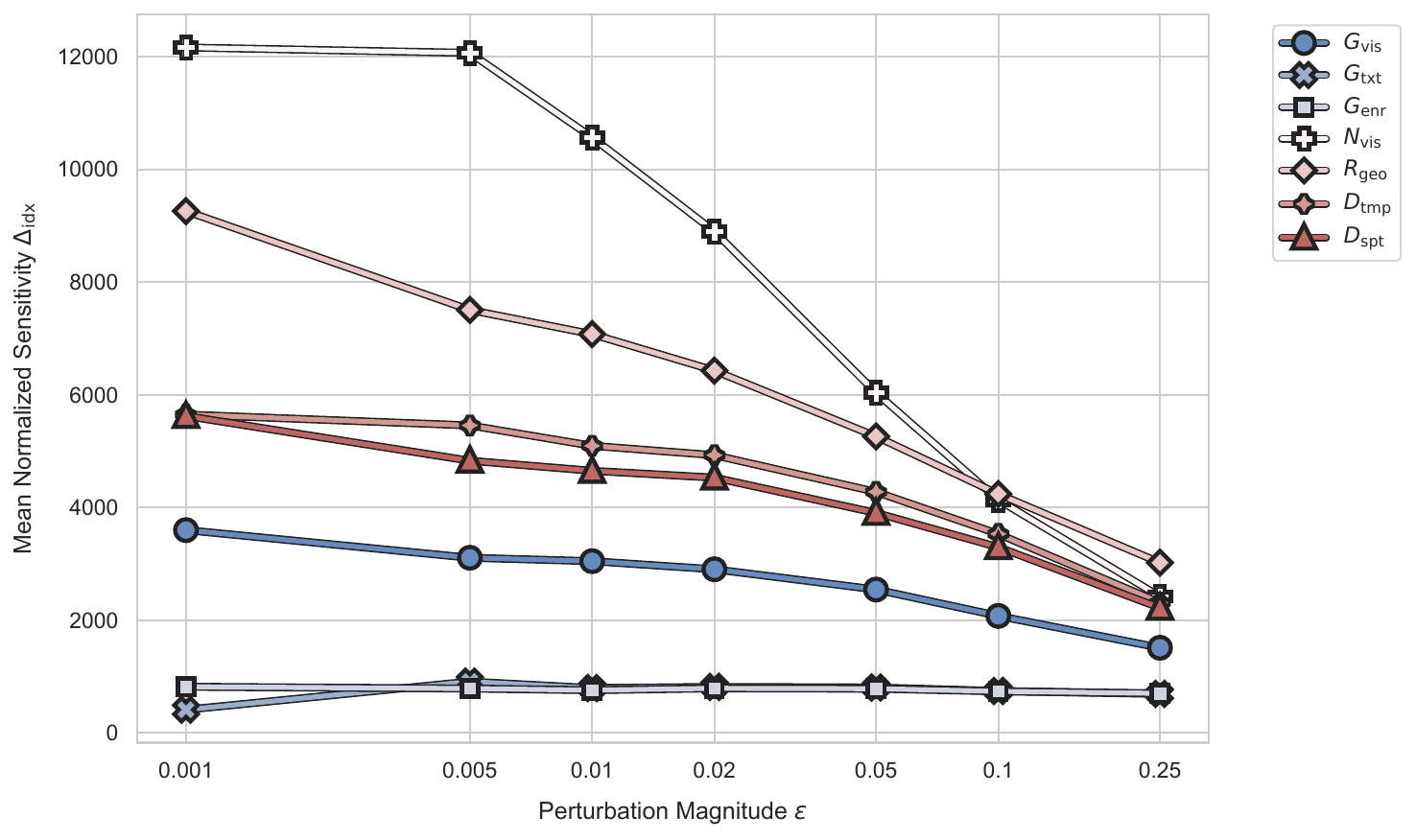}
        \caption{
            \textbf{Mean normalized sensitivity in relation to perturbation
            magnitude}.
            For each proposed subscore weight and perturbation magnitude
            $\epsilon$, the normalized output sensitivity is computed as the mean
            output distance $\Delta_{\text{idx}}$ across $n=10{,}000$ sampled base
            weight configurations, divided by $\epsilon$.
            The x-axis is displayed on a logarithmic scale.}
        \label{fig:fig3}
    \end{figure}

    Fig.~\ref{fig:fig3} displays the mean normalized sensitivity over
    different $\epsilon$ measured using $\Delta_{\text{idx}}$.
    The sensitivity ranking of the subscore weights remains stable over the
    different values of $\epsilon$.
    While the normalized sensitivity subsides for $N_{\text{vis}}$ and
    $R_{\text{geo}}$ with increasing $\epsilon$, $G_{\text{txt}}$ and
    $G_{\text{enr}}$ remain stable over the examined range, with exceptions
    limited to the most extreme values of $\epsilon=0.001$ and $0.25$.
    To contextualize the magnitude of the observed frame-selection changes, we
    computed an empirical random-selection baseline by comparing each of the
    $10,000$ unperturbed selections with $1{,}000$ randomly sampled sets of
    $K=5$ frames from the same image sequence.
    The resulting mean unnormalized index distance $\Delta_{\text{idx}}$ was
    $1316.4$ frames.
    In comparison, the mean unnormalized distances induced by the evaluated
    perturbation magnitudes
    $\epsilon \in \{0.001, 0.005, 0.01, 0.02, 0.05, 0.1, 0.25\}$ ranged from
    $0.4\text{--}12.2$, $3.9\text{--}60.3$, $7.6\text{--}105.7$,
    $15.8\text{--}177.9$, $39.1\text{--}301.9$, $73.5\text{--}423.7$,
    $172.4\text{--}754.6$ frames for each $\epsilon$, respectively, across the
    seven weights, and remained below the random-selection baseline.

    For the evaluated dataset, the analysis showed highly consistent behavior
    across the three sensitivity metrics $\Delta_{\text{idx}}$,
    $\Delta_{\text{tmp}}$ and $\Delta_{\text{spt}}$, with the spatial and temporal
    metrics yielding the same qualitative conclusions as the index metric.
    The mean normalized temporal and spatial sensitivities exhibited Spearman
    correlations of \(\rho=0.99\) and \(\rho=0.98\), respectively, with the
    corresponding index-based sensitivity across the evaluated weights and
    perturbation magnitudes.
    We therefore report only the index metric $\Delta_{\text{idx}}$ in the main
    paper, while providing the corresponding spatial sensitivity results and
    data in the
    \href{https://datahub.rz.rptu.de/hstr-csrl-public/publications/multimodal-spatio-temporal-summarization}
    {external repository}.

    Fig.~\ref{fig:fig4} illustrates five representative summaries using the
    proposed pipeline applied to the evaluation dataset.

    All raw and intermediate evaluation data, the code used to generate it,
    and the used seeds to reproduce the reported results, are provided in the
    \href{https://datahub.rz.rptu.de/hstr-csrl-public/publications/multimodal-spatio-temporal-summarization}
    {external repository}.
    The raw dataset itself is the only data not provided due to privacy
    concerns; see Section~\nameref{sec:data_availability} for details.

    \definecolor{summaryRed}{HTML}{ff5555}
    \definecolor{summaryBlue}{HTML}{2a7fff}
    \definecolor{summaryYellow}{HTML}{ffdd55}
    \definecolor{summaryGreen}{HTML}{5aa02c}
    \definecolor{summaryTeal}{HTML}{80e5ff}

    \begin{figure}[!htbp]
        \centering
        \includegraphics[width=0.95\columnwidth]{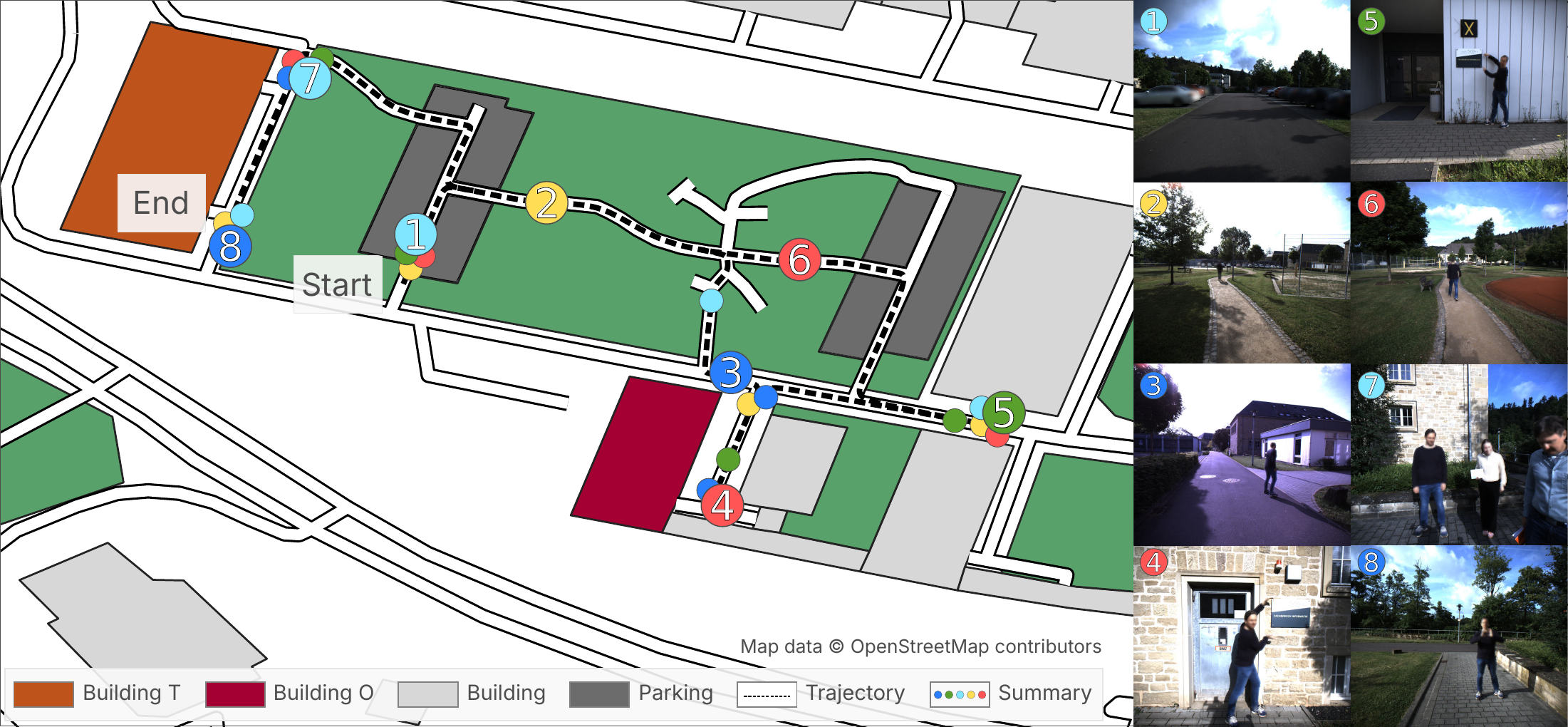}
        \caption{
            \textbf{Exemplary visual summaries on the evaluation dataset}.
            The figure illustrates five representative summaries with
            $\text{top-K} = 5$, generated using distinct weight configurations:
                {$\color{summaryBlue}\bullet$} $w_{G\text{vis}} = 0.3, w_{R\text{geo}} = 0.7$,
                {$\color{summaryGreen}\bullet$} $w_{G\text{txt}} = 1.0$,
                {$\color{summaryTeal}\bullet$} $w_{G\text{enr}} = 1.0$,
                {$\color{summaryYellow}\bullet$} $w_{D\text{tmp}} = 1.0$,
                {$\color{summaryRed}\bullet$} $w_{D\text{spt}} = 1.0$ with all
            remaining weights set to 0.
            Each colored point represents the agent's location at the time a
            selected image was recorded.
            For $R_{\text{geo}}$, the T- and O-Buildings serve as the OSM
            target objects.
        }
        \label{fig:fig4}
    \end{figure}

    \subsection{Discussion}\label{subsec:discussion}
    As previously stated, $G_{\text{txt}}$ and $G_{\text{enr}}$ display the
    largest correlation between their local sensitivities.
    Importantly, this observation concerns the sensitivities of the frame
    selection process and does not establish a correlation between the
    subscores themselves.
    One possible interpretation is that the two subscores may partially overlap
    in their influence on frame selection.
    This hypothesis is plausible given their construction, as the descriptions
    used for $G_{\text{enr}}$ are generated from those used for
    $G_{\text{txt}}$.
    However, establishing complementarity or redundancy between the subscores
    would require a direct analysis of the subscores themselves.
    Finally, the correlation patterns remain consistent across the examined
    perturbation magnitudes $\epsilon$, suggesting that the observed
    sensitivity correlations are not artifacts of a particular choice of $\epsilon$.
    For $\epsilon = 0.25$, this pattern exhibits notable exceptions.
    Among others, the previously observed correlation involving the local
    sensitivities of $G_{\text{vis}}$ and $D_{\text{spt}}$ deviates from those
    observed at smaller perturbation magnitudes.
    This can plausibly be attributed to the comparatively large
    $\epsilon = 0.25$.
    Since the expected value of each randomly sampled simplex weight is
    $\mathbb{E}[w_i] = \frac{1}{7} \approx 0.14$, the applied perturbation is,
    on average, larger than the original weight itself.
    Thus, $\epsilon = 0.25$ no longer represents a local perturbation.

    Given the mean normalized sensitivity, the weight associated with subscore
    $N_{\text{vis}}$ is the one the frame selection is most sensitive to.
    This ranking holds true for most values of $\epsilon$, with again the
    exception of $\epsilon = 0.25$, see above.
    Both $G_{\text{txt}}$ and $G_{\text{enr}}$ show the lowest mean normalized
    sensitivity and their normalized sensitivity remains the most stable across
    different values of $\epsilon$.
    Both subscores assign identical values to 60 consecutive frames, reducing
    their ability to distinguish between neighboring frames.
    This batching behavior may contribute to the comparatively low local
    sensitivity observed for their associated weights.

    The near-perfect correlations between the sensitivities measured by
    \(\Delta_{\mathrm{idx}}\) and \(\Delta_{\mathrm{tmp}}\) (\(\rho=0.99\)),
    and between \(\Delta_{\mathrm{idx}}\) and \(\Delta_{\mathrm{spt}}\)
    (\(\rho=0.98\)), may be attributed to two factors.
    First, the index- and time-based metrics capture closely related temporal
    displacements, with differences arising primarily from the non-uniform
    sampling of the recorded image frames due to the characteristics of the
    image-capturing software and hardware.
    Second, the approximately constant traversal speed observed in the present
    recording results in a strong relationship between temporal and spatial
    displacement, thereby yielding similar sensitivity trends across the three
    metrics.

    The provided exemplary visual summaries in Fig.~\ref{fig:fig4} illustrate
    the behavior of the pipeline on the evaluation dataset.
    The weight configuration $w_{G\text{vis}}=0.3$ and
    $w_{R\text{geo}}=0.7$, with T- and O-Buildings as the \gls*{osm} target
    objects, successfully clusters the selected images around these buildings.
    In particular, the contribution of $w_{G\text{vis}}$ promotes visual
    diversity, while $w_{R\text{geo}}$ primarily governs the spatial
    relevance of the selected images, resulting in a combination of diverse
    views centered around the target buildings.
    The pronounced clustering of selected images from different weight
    configurations at the extremes of the trajectory, such as the upper-left
    region, the far right, and the starting point, may be attributed to several
    factors:
    First, because the gain-based and distribution subscores are constant
    across candidates when $|\mathcal{S}|=0$, configurations dominated by these
    terms can yield identical combined scores in the first iteration.
    The specified earliest-timestamp tie-breaking rule then selects the first
    frame, producing the observed effect.
    Specifically, the first frame was present in 0.99\% of image sets
    (99~of~10,000) under sampled weights and 0.99\% (4,862~of~490,000) under
    perturbed weights.
    This is compared to an expected baseline of $\frac{5}{8464} \approx 0.06\%$
    for a uniform distribution across the 8,464 images in the recording.
    This effect is primarily confined to configurations where gain-based
    and spatio-temporal distance subscores dominate while non-gain metrics
    remain zero.
    Thus, in the given examples in Fig.~\ref{fig:fig4} the effect is
    particularly pronounced, since most given configurations contain only one
    non-zero subscore weight.
    Second, the circular structure of the recorded trajectory creates a
    spatio-temporal correspondence, causing the image sets selected based on
    spatial and temporal distribution to be correlated within this dataset.
    Third, several turning points and endpoints along the trajectory coincide
    with notable actions of the recorded protagonist, further linking subscores
    such as $G_{\text{vis}}$ to the underlying spatio-temporal structure.
    This effect may also contribute to the observed correlations between the
    location-related sensitivities, warranting further investigation across
    more diverse datasets.

    \section{Conclusion and Future Work}\label{sec:conclusion-and-future-work}
    We introduced a pipeline for selecting representative frames from
    localized image sequences, providing users with explicit control over the
    criteria governing the resulting visual summary.
    Our evaluation demonstrated the principal capabilities of the approach and
    showed that different weight configurations can affect frame selection.
    The sensitivity analysis further revealed consistent patterns in the
    relative influence of the individual weights across perturbation
    magnitudes.

    However, the current evaluation is limited to a single dataset and a single
    trajectory, restricting the extent to which the observed weighting
    and sensitivity patterns can be generalized to other trajectories or
    application contexts.
    Future work will therefore evaluate the pipeline across more diverse
    datasets, trajectories, and domains.

    \section*{Data Availability and Protection}\label{sec:data_availability}
    The data used for this paper were recorded at the campus of Trier University
    of Applied Sciences.
    The raw data are confidential and only shared with reviewers upon request,
    with personal data being anonymized.
    All published data are anonymized and re-identified persons are
    pseudonymized.
    All recorded, confidential and published data are exclusively stored on
    servers at Trier University of Applied Sciences and RPTU University
    Kaiserslautern-Landau.
    Code and results are available at
    \href{https://datahub.rz.rptu.de/hstr-csrl-public/publications/multimodal-spatio-temporal-summarization}
    {https://datahub.rz.rptu.de/hstr-csrl-public/publications/multimodal-spatio-temporal-summarization}.

    \section*{Acknowledgments}\label{sec:acknowledgments}
    The authors used ChatGPT (OpenAI) to assist in generating portions of the
    software code and to improve grammar and clarity throughout the manuscript.
    All AI-assisted content was reviewed and validated by the authors, who take
    full responsibility for the final work.

    \bibliographystyle{IEEEtran}
    \bibliography{literature}

\end{document}